\documentclass[journal]{IEEEtran}

\ifCLASSINFOpdf
\else
   \usepackage[dvips]{graphicx}
\fi
\usepackage{url}
\usepackage{graphicx}
\usepackage{comment}
\usepackage{amsmath,amssymb} % define this before the line numbering.
\usepackage{color}
\usepackage{caption}
\usepackage{diagbox}
\usepackage{stfloats}
\usepackage{float}
\usepackage{cite}

\usepackage{graphicx}
\usepackage{float}
\usepackage{color, soul}
\usepackage{authblk}
\usepackage{hyperref}
\title{More than one Author with different Affiliations}
\author[a]{Xiaofeng Zhang}
\author[b]{Feng Chen}
\author[a]{Cailing Wang \thanks{The first authors are Xiaofeng Zhang and Feng Chen. Corresponding authoris Cailing Wang and her email is wangcl@njupt.edu.cn}}
\author[a]{Ming Tao}
\author[a]{Songsong Wu}
\author[a]{Guoping Jiang}

\affil[a]{College of Automation, Nanjing University of Posts and Telecommunications, Nanjing, China}
\affil[b]{School of Computer Science, Nanjing University of Posts and Telecommunications, Nanjing, China}

\date{}

\begin{document}

\title{SiENet: Siamese Expansion Network for Image Extrapolation}

\markboth{Journal of \LaTeX\ Class Files, Vol. 14, No. 8, August 2020}
{Shell \MakeLowercase{\textit{et al.}}: Bare Demo of IEEEtran.cls for IEEE Journals}

\maketitle
\begin{abstract}
Different from image inpainting, image outpainting has relatively less context in the image center to capture and more content at the image border to predict. Therefore, classical encoder-decoder pipeline of existing methods may not predict the outstretched unknown content perfectly. In this paper, a novel two-stage siamese adversarial model for image extrapolation, named Siamese Expansion Network (SiENet) is proposed. Specifically, in two stages, a novel border sensitive convolution named adaptive filling convolution is designed for allowing encoder to predict the unknown content, alleviating the burden of decoder. Besides, to introduce prior knowledge to network and reinforce the inferring ability of encoder, siamese adversarial mechanism is designed to enable our network to model the distribution of covered long range feature as that of uncovered image feature. The results on four datasets has demonstrated that our method outperforms existing state-of-the-arts and could produce realistic results. Our code is released on \url{https://github.com/nanjingxiaobawang/SieNet-Image-extrapolation}.
\end{abstract}


\IEEEpeerreviewmaketitle

\section{Introduction}

\begin{figure}[t]
\begin{center}
\includegraphics[width=0.5\textwidth]{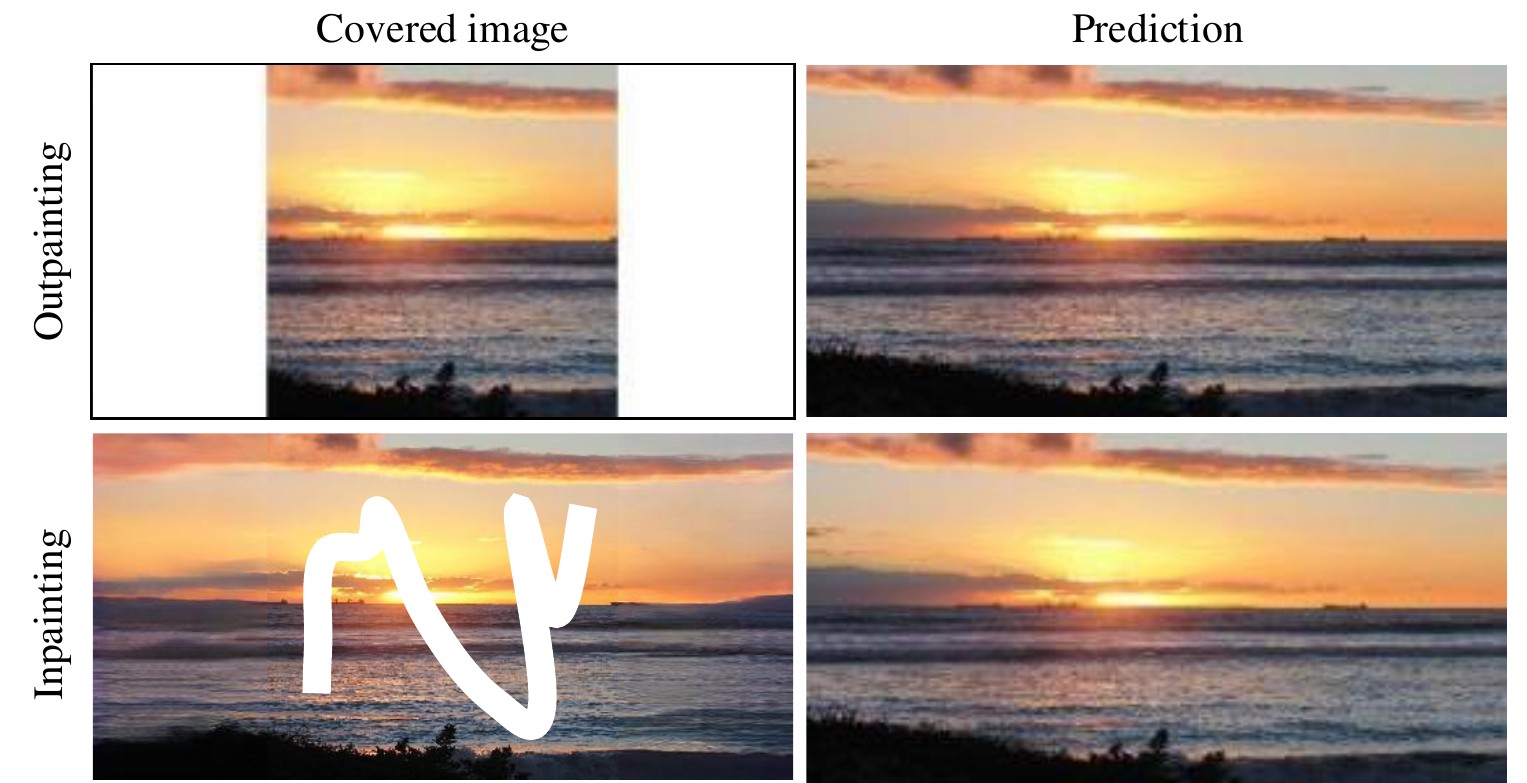}
\end{center}
  \caption{Overview of difference between image inpainting and image outpainting.}
\label{Figure:task_difference}
\end{figure}

Image extrapolation, as illustrated in Fig. \ref{Figure:task_difference}, is to generate new contents beyond the original boundaries of a given image.
Even belonging to the general image painting task as inpainting \cite{Patchmatch,context-encoder ,High-resolution,generative-attention,image-inpainting1,image-inpainting2} , image outpainting \cite{photo-realistic} has its special characteristics. We rethink this task from two aspects. First, comparing to inpainting, image extrapolation would rely on relative less context to infer much larger unknown content. Besides, the inferred content locates outside the given image. Therefore, existing expertise of inpainting can not be applied into this task directly. Second, existing methods of image extrapolation generally employ classical encoder-decoder (ED) structure. This structure forces the encoder to capture the global and local features and allows decoder to recover them to desired resolution. Namely, all the burden of
inferring is put on the decoder. Yet, the input of outpainting doesn't have such abundant features, which makes the capturing ability of encoder weak and the inferring burden of decoder heavy.

Due to the requirement of outpainting task, most inpainting expertise can't be applied to it directly. \cite{Outpainting-srn} proposed the specific spatial expansion module and boundary reasoning module for the requirement of connecting marginal unknown content and inner context. Besides, heavy burden of prediction on decoder may result in the poor performance of generating realistic images. \cite{image-outpainting} applied DCGAN  \cite{DCGAN} which mainly focuses on the prediction ability of decoder. This kind of methods could predict the missing parts on both sides of the image, while this design would ignore coherence of semantic and content information.

\begin{figure*}[t]
\begin{center}
\includegraphics[width=1\textwidth]{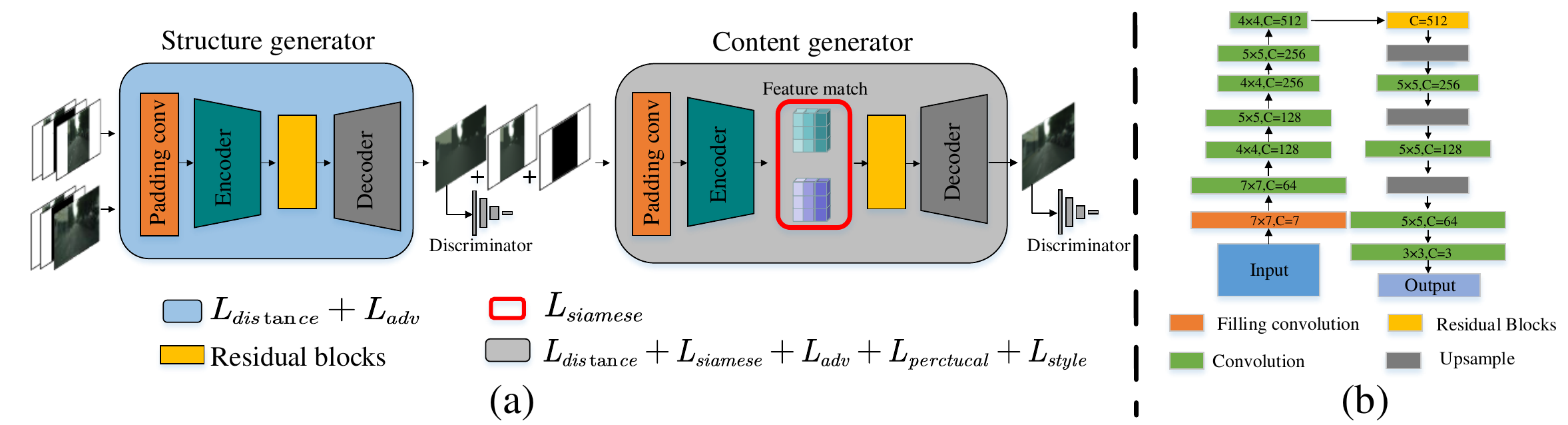}
\end{center}
  \caption{Illustration of network architecture. (a) is the basic pipeline of our network. (b) is the detailed structure of structure generator, which is similar to that of content generator.}
\label{Figure:pipeline}
\end{figure*}

In this paper, to design task-specific structure, we propose a novel two-stage siamese adversarial model for image extrapolation, named Siamese Expansion Network (SiENet). In our SiENet, a boundary-sensitive convolution, named adaptive filling convolution, is proposed to automatically infer the features of surrounding pixels outside known content with balance of smoothness and characteristic. This adaptive filling convolution is inserted to the encoder of two-stage network, activating the sensitivity of encoder for border features. Therefore, encoder could infer the unknown content and the inferring burden of decoder could be alleviated.

In addition, the siamese adversarial mechanism is designed to introduce prior knowledge into network and adjust the inferring burden of each part. In the joint training of two stages, ground truth and covered image are fed into network to calculate the siamese loss. Siamese loss encourages the features of them in the subspace to be similar, leading to reinforced predicting ability of long range encoder. Besides, a adversarial discriminator is designed in each stage to push the global generator to generate realistic prediction. Thus, the whole inferring burden of the network is reasonably allocated.

Our contributions can be summarized as follows:

\begin{itemize}
\item We design a novel two-stage siamese adversarial network for image extrapolation. Our SiENet, a task-specific pipeline, could regulate the inferring burden of each part and introduce prior knowledge into network legitimately.
\item We propose an adaptive filling convolution to concentrate on inferring pixels in unknown area. By inserting this convolution into encoder, encoder could be endowed powerful inferring ability.
\item Our method achieves promising performance on four datasets and outperforms existing state-of-the-arts. The results on four datasets, i.e., Cityscapes, paris street-view, beach and Scenery, including street and nature cases, indicate the robustness of our method.
\end{itemize}

\section{Methodology}

Our SiENet is a two-stage network, and each stage has specific generator and discriminator, as shown in Fig. \ref{Figure:pipeline} (a). The details of our method are provided as follows.

\subsection{Framework Design}
Given an input image $X\in \mathbb{R}^{{h}\times w\times c}$, extension filling map $M\in \mathbb{R}^{{h}\times w\times 1}$ and smooth structure $S\in \mathbb{R}^{{h}\times w\times c}$ \cite{structure-extraction}, our method intends to generate a visually convincing image $\widehat{Y}\in \mathbb{R}^{h \times w^{'}\times c}$ where $w^{'}$ is desired width of the generated image. SieNet follows the coarse-to-fine two-stage pipeline \cite{Structure-flow}. As shown in Fig. \ref{Figure:pipeline} (a), an uncovered smooth structure $S_{gen}\in \mathbb{R}^{{h}\times w^{'}\times c}$ is produced by the structure generator first. Then the generated structure feature $S_{gen}$ would combine with image $X$ and filling map $M$, acting as the input of second generator, i.e., content generator, to generate outstretched image $\widehat{Y}$. Besides, the discriminator $D$ to ensure the output of generator and actual data conform to a certain distribution.

As illustrated in Fig. \ref{Figure:pipeline} (b), the encoder of structure generator downsamples the input with 8$\times$ scale. Then to capture multi-scale information two residual blocks are designed to further proceed the output of encoder. Finally, by three nearest-neighbor interpolation, the output of residual blocks is upsampled to desired resolution. The detailed structure of content generator is similar to that of structure generator, except adding two residual blocks before last two upsampling operations. The discriminator $D$ follows the structure and protocol of that of BicycleGAN \cite{BicycleGAN,Global-and-local} .

\begin{figure}[t]
\begin{center}
\includegraphics[width=.38\textwidth]{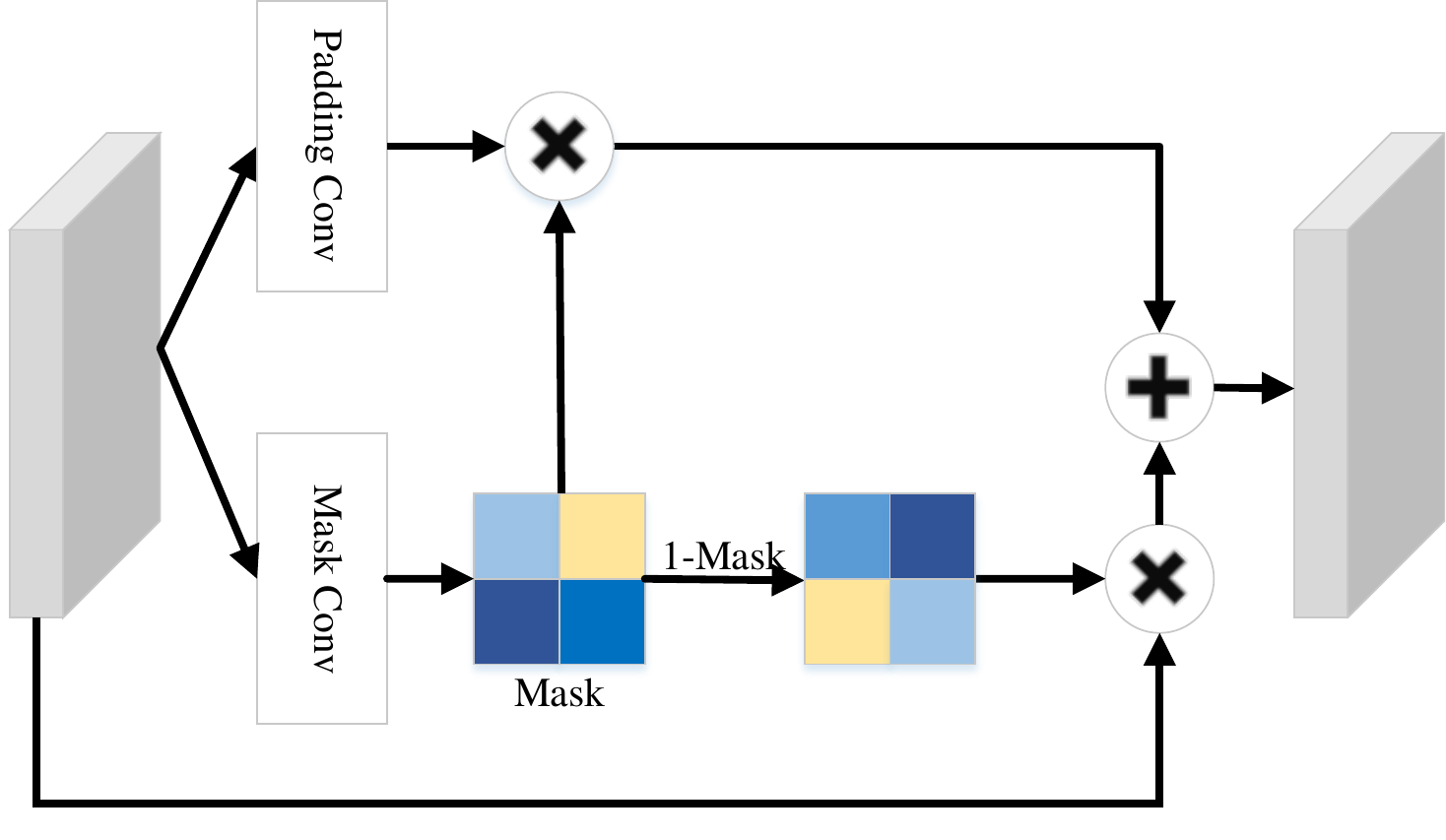}
\end{center}
  \caption{Scheme of adaptive filling convolution.}
\label{Figure:scheme}
\end{figure}

\subsection{Adaptive Filling Convolution}

To endue encoder with predicting ability, we propose a boundary sensitive convolution, i.e., adaptive filling convolution. Given a convolutional kernel with $K$ sampling locations (e.g., $K=9$ in a $3\times 3$ convolution), let $w_k$ and $m_k\in [0,1]$ denote the weight of padding convolution and mask modulation for $k$-th position. $x(p)$ and $y(p)$ represent the feature map at location $p$ from input and output respectively. Therefore, the filling convolution could be formulated as follows:
\begin{equation}\label{padding_conv}
  y(p_0) = \sum_{k=1}^{K}{[w_k \cdot m_k \cdot x(p_0 + p_k)+ (1-m_k)\cdot x(p_0+p_k)]}
\end{equation}

In practical learning , the scheme of our filling convolution is shown as Fig. \ref{Figure:scheme}. Two separate convolutions model padding weight $w_k$ and mask modulation $m_k$ respectively where $m_k$ is defaulted to 1. Our filling convolution could predict the pixel outside the image boundary. When kernel center locates just outside known content, only scant real pixels would join the patch calculation. Our padding convolution would predict it by known information with $w_k$, and another skip connection from input to output would keep characteristic value to avoid high variance. Besides, mask modulation $m_k$ works as a adaptive balance factor of these two branches, to guarantee the smoothness and characteristic of prediction. We insert a $7\times 7$ filling convolution into the bottleneck of encoder, therefore, encoder could possess the ability of capturing context and the ability of predicting unknown content.

\subsection{Siamese Adversarial Mechanism}
Classical adversarial mechanism could constrain the whole generator toward the real case. However, this constraint is implicit to each part of generator. Especially in two-stage GAN \cite{GAN,WGAN,Impoved-WGAN} for image outpainting, long range encoding may lead to insufficient inferring ability for decoder. Considering the predicting ability of encoder brought by filling convolution, we further add explicit constraint in encoder to push the features of covered and uncovered image to be common. Namely, prior knowledge of uncovered image is learned by encoder. In addition, we also take advantages of adversarial mechanism in our model to constrain the generated results to be consistent with ground truth \cite{Siamese1,Siamese2}. Through these two thoughtful constraints brought by our siamese adversarial mechanism, the predicting burden is well regulated.

Let $G_{strut}$ denotes structure generator and $E_{cont}$ denotes encoder of content generator, $I= [X,M,S]$ is the input of the whole network. The output $F$ of long range encoder $E_{cont}(G_{strut}())$ could be formulated as:

\begin{equation}\label{long_range_encoder}
  F = E_{cont}(concat(X,M,G_{strut}(I)))
\end{equation}

Therefore, let superscripts $^{'}$ and $^{gt}$ denote whom is generated by covered input and ground truth input respectively, the siamese loss $L_{siamese}$ is ${\ell 2}$ loss of $F^{'}$ and $F^{gt}$:
\begin{equation}\label{siamese_loss}
L_{siamese} = ||F^{'}-F^{gt}||_2
\end{equation}

In order to generate realistic results, we introduce adversarial loss $L_{adv}$ in the structure generator:
\begin{equation}
\begin{split}
L_{adv}=\mathbb{E}[\log (1-D(\widehat{Y})]+ \mathbb{E}\left[\log D\left(Y\right)\right]
\end{split}
\end{equation}

\subsection{Loss Function}

We introduce $L_{distance}$ loss to predict the distance between the structures $S_{gen}^{'}$ and $S^{gt}$:
\begin{equation}
L_{distance}=\left\|S_{gen}^{'}-S^{gt}\right\|_{1}
\end{equation}

Besides, the perceptual loss $L_{perctual}$ and the style loss $L_{style}$ \cite{Image-style} are applied into our network. $L_{perctual}$ is defined as:
\begin{equation}
L_{perctual}=\mathbb{E}\left[\sum_{i} \frac{1}{N_{i}}\left\|\phi_{i}\left(Y\right)-\phi_{i}\left(\widehat{Y}\right)\right\|_{1}\right],
\end{equation}
$\phi_{i}$ is the activation map of the $i$-th layer of the pre-trained network. In our work, $\phi_{i}$ is activation map from the layers of relu1-1, relu2-1, relu3-1, relu4-1 and relu5-1 of the ImageNet pre-trained VGG-19. These activation maps are also used to calculate the style loss to measure the covariance between the activation maps. Given a feature map size C$_{j}$ $\ast$ H$_{j}$ $\ast$ W$_{j}$, the style loss $L_{style}$ is calculated as follows.

\begin{equation}
L_{style}=\mathbb{E}\left[\left\|G^{\phi}\left(\widehat{Y}\right)-G^{\phi}\left(Y \right)\right\|_{1}\right],
\end{equation}
$G^{\phi}$ is the the reconstructed Gram metric based on activation map $\phi_j$. Totally, the overall loss $L_{total}$ is:
\begin{equation}
\begin{split}
L_{total}=& \lambda_{dist} L_{distance}+\lambda_{adv} L_{adv}+\lambda_{p} L_{perctual}\\
&+ \lambda_{s} L_{style} + \lambda_{sie}L_{siamese},
\end{split}
\end{equation}
where $\lambda_{dist}$, $\lambda_{adv}$, $\lambda_{p}$, $\lambda_{s}$ and $\lambda_{sie}$ are set 5, 1, 0.1, 250 and 1 respectively.

\section{Experiments}
\subsection{Implementation Details and Configurations}
 Our method is flexible for two-direction and single-direction outpainting. For two-direction outpainting, our method is evaluated on three datasets, i.e., Cityscapes \cite{cityscapes}, paris street-view \cite{context-encoder} and beach \cite{image-outpainting}. Single-direction evaluation is made on Scenery dataset \cite{NS-outpainting}. Besides, we take structural similarity (SSIM), peak signal-to-noise ratio (PSNR) as our evaluation metrics.

In training and inference, the images are resized to $256\times 256$. We train our model using Adam optimizer \cite{Adam} with learning rate $0.0001$, beta1 $0$ and beta2 $0.999$. The batch size is set to 8 for paris-street view and beach, 2 for Cityscapes, and 16 for Scenery. The total iteration of joint training is $10^6$ for four datasets.

\begin{figure}[b]
\begin{center}
\includegraphics[width=.5\textwidth]{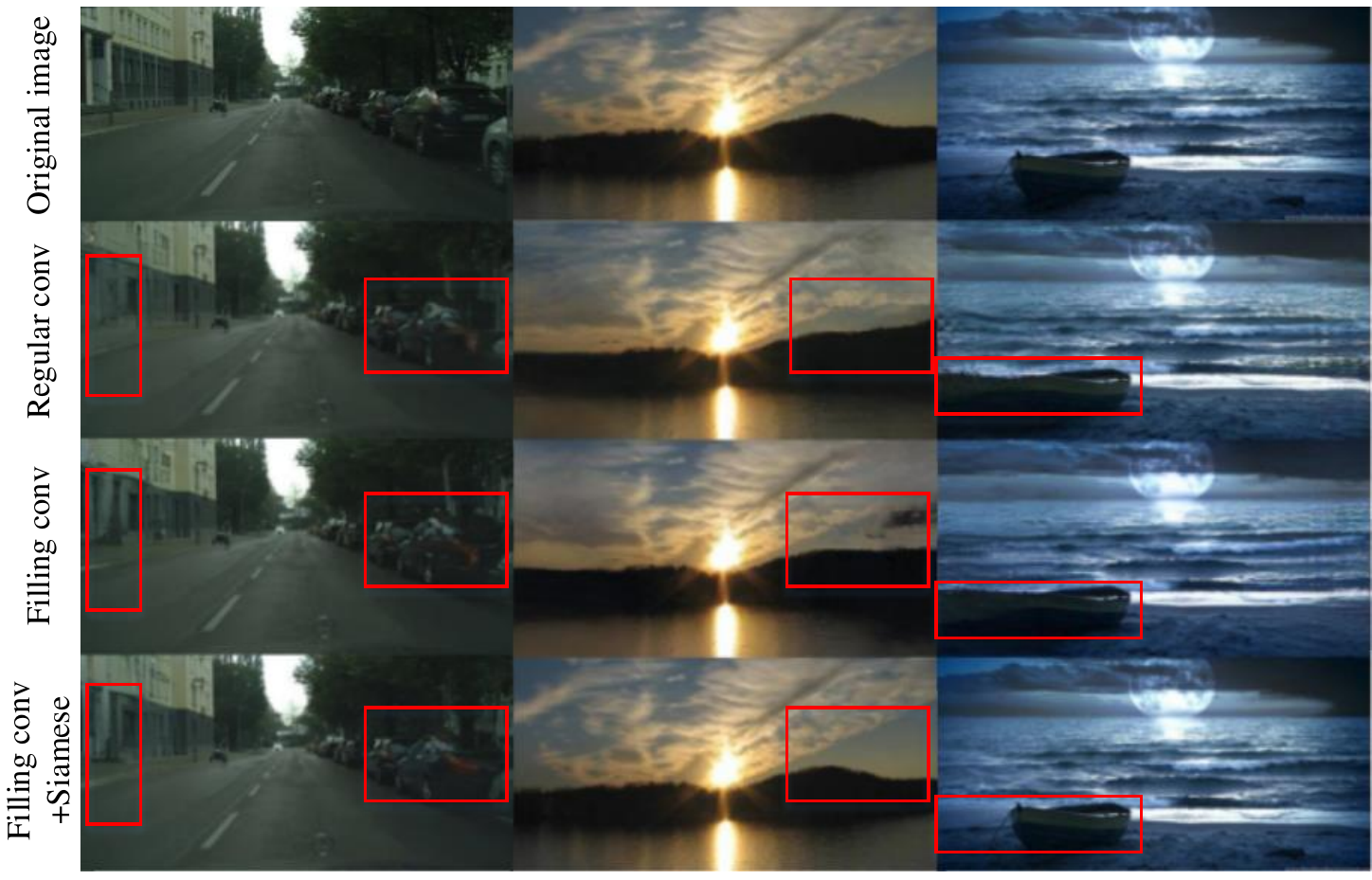}
\end{center}
  \caption{Ablation study of effectiveness of filling convolution and siamese adversarial mechanism on Cityscapes and beach.}
\label{Figure:ablation}
\end{figure}

\begin{table}[h]\caption{Ablation experiments of filling convolution and siamese adversarial mechanism on Cityscapes. `FConv' and `SAM' denote filling convolution and siamese adversarial mechanism respectively.}
\centering
\begin{tabular}{c|c|c|c}
\hline
FConv & SAM  & SSIM & PSNR \\
\hline
&     &     0.6896    & 23.2539            \\
  $\surd$    &     &    0.7254     &23.5987             \\
      & $\surd$    &    0.7647     &24.4578                \\
  $\surd$    & $\surd$           &  0.7832    &  24.8213        \\
      \hline
\end{tabular}\label{Table:ablation}
\end{table}

\subsection{Ablation Study on SiENet}

\begin{table*}[t]
  \begin{center}
    \caption{Performance comparison of image two-direction and single-direction outpainting. The two-direction outpainting is evaluated on beach, paris-street view and Cityscapes, and single-direction outpainting is evaluated on Scenery.}
  \small
    \begin{tabular}{||c||c|c|c|c||c|c|c|c||}
\hline \multicolumn{1}{|c||} {} & \multicolumn{4}{c||} {\text { SSIM }} & \multicolumn{4}{c||} {\text { PSNR }}  \\
\hline \text {\diagbox{Methods}{Datasets}}  & Beach & Paris & Cityscapes&Scenery  & Beach & Paris & Cityscapes  &Scenery \\
\hline \text { Image-Outpainting } &0.3385 &0.6312  & 0.7135 &- & 14.6256 & 19.7400 & 23.0321&- \\
\hline \text { Outpainting-srn } & 0.5137 & 0.6401 & 0.7764 &- & 18.2211 &19.2734  & 23.5927&- \\
\hline \text{Edge-Connect}& 0.6373 & 0.6342 & 0.7454 &- & 19.8372 & 22.1736& 24.1413&- \\
\hline \text{NS-Outpainting}&-& -&-& 0.6763&-&-&-&17.0267\\
\hline \text { Our method } & $\mathbf{0.6463}$ &$\mathbf {0.6428}$ & $\mathbf{0.7832}$ &$\mathbf{0.8557}$ &$\mathbf{20.7965}$ & $\mathbf{23.9794}$ & $\mathbf{24.8213}$ &$\mathbf{31.7686}$ \\
\hline
\end{tabular}
\end{center}
\end{table*}\label{Table:results}

\begin{figure*}[t]
\begin{center}
\includegraphics[width=1\textwidth]{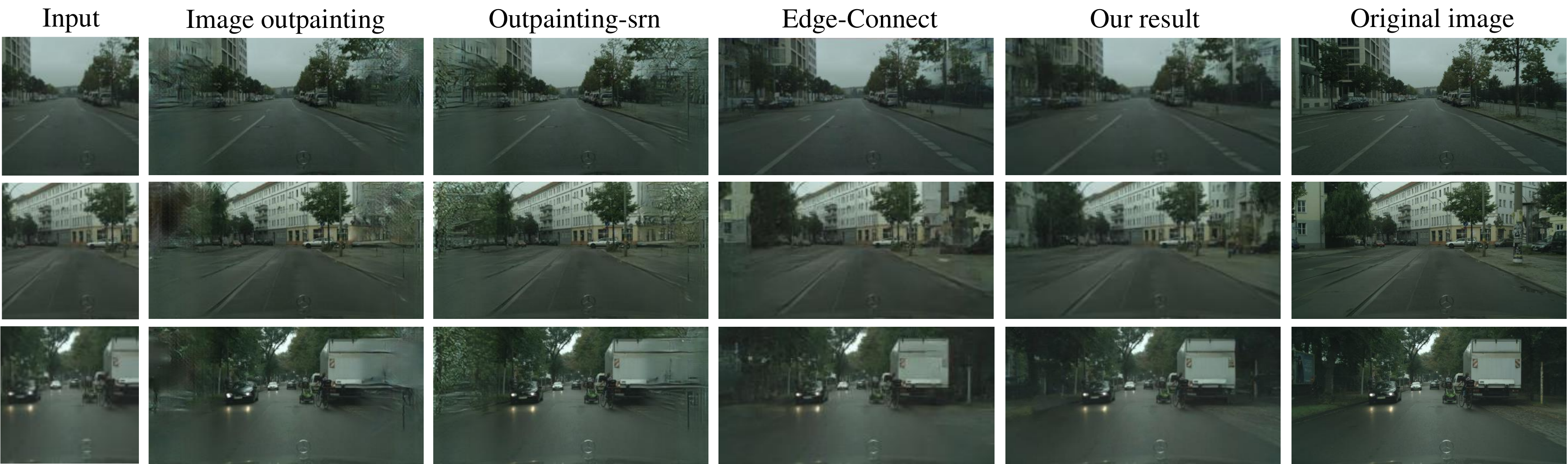}
\end{center}
  \caption{Quantitative two-direction comparisons of our method with our state-of-the-arts on Cityscapes.}
\label{Figure:comparison}
\end{figure*}

\subsection{Comparisons}

\begin{figure}[t]
\begin{center}
\includegraphics[width=.5\textwidth]{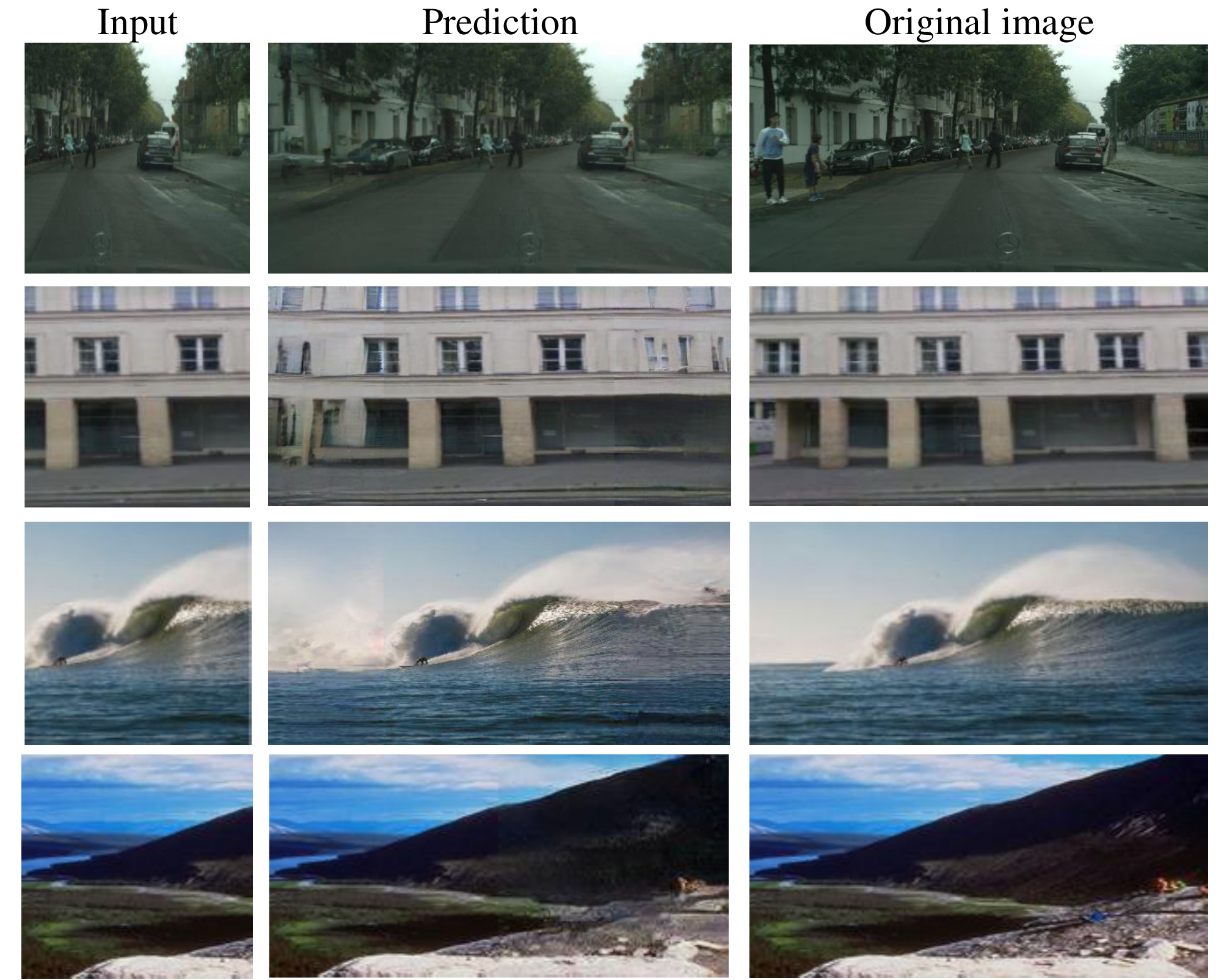}
\end{center}
  \caption{The visual performance of our method on four dataset. The front three rows represent two-direction outpainting results. The last row is single-direction outpainting result.}
\label{Figure:results}
\end{figure}

To validate the effectiveness of filling convolution and siamese adversarial mechanism in our network, We make the ablation experiment in Tab. \ref{Table:ablation}. Only using filling convolution or  siamese adversarial
mechanism could get a significant improvement over classical encoder-decoder. And the combination of them could achieve the best performance, indicating the regulation of predicting burden is beneficial for the task. We also visualize the results of street and nature scenery to measure the quantitative capability of them. As shown in Fig. \ref{Figure:ablation}, the combination of these proposed two parts could recover the details and generate realistic results in easy nature and complicated street cases.

Our method is compared with existing methods including Image-Outpainting \cite{image-outpainting}, Outpainting-srn \cite{Outpainting-srn}, Edge-connect \cite{edge-connect} and NS-outpainting  \cite{NS-outpainting}. As shown in Tab. \uppercase\expandafter{\romannumeral2}, for two-direction outpainting, our method outperforms Image-Outpainting \cite{image-outpainting}, Outpainting-srn \cite{Outpainting-srn}, Edge-connect \cite{edge-connect}, and achieves state-of-the-art performance in three datasets. For single-direction, the great margin of performance between our method and NS-outpainting  \cite{NS-outpainting} indicates the superiority and generality of our SiENet. Besides, we also provide visual comparison of various methods in Fig. \ref{Figure:comparison} and visual performance of our method on four datasets in Fig \ref{Figure:results}. Notably, the images generated by our method are similar to the ground truth than that of existing methods. 

% \begin{table}[h]
%   \begin{center}
%     \caption{Performance comparison of single-direction outpainting on Scenery dataset.}
%   \small
%     \begin{tabular}{c|cc}
%       \hline
%       Method& SSIM &PSNR \\
%       \hline
%       NS-outpainting &0.6763&	17.0267 \\
%       \hline
%       Our method &0.8557	&31.7686\\
%       \hline
%     \end{tabular}
%   \end{center}
% \end{table}\label{Table:result_scenry}

\section{Conclusion}
 A novel end-to-end model named Siamese Expansion Network for Image Extrapolation is propoesd in this paper. To regulate the heavy predicting burden of decoder legitimately, we first propose adaptive filling convolution to endow encoder with predicting ability. Then we introduce siamese adversarial mechanism in long range encoder to allow encoder to learn prior knowledge of real image and reinforce the inferring ability of encoder. Our method has achieved promising performance on four datasets and outperforms existing state-of-the-arts.

\section*{Acknowledgement}
The work described in this paper was fully supported by the National Natural Science Foundation of China project (61871445) and  Nanjing University of Posts and Telecommunications General School Project (NY22057).

% \bibliographystyle{plain}
% \bibliography{SPL}

\end{document}